\documentclass[11pt,a4paper]{article}
\usepackage[hyperref]{acl2018}
\usepackage{times}
\usepackage{latexsym}
\usepackage{amssymb,amsmath}

\usepackage{url}

\usepackage{enumitem}
\usepackage{booktabs}

\aclfinalcopy 

\newcommand{\seq}{\textsc{Seq2Seq}\ }

\newcommand{\R}{\mathbb{R}}

\title{Automatic Evaluation of Neural Personality-based Chatbots}

\author{Yujie Xing \and Raquel Fern\'andez\\
Institute for Logic, Language and Computation\\
University of Amsterdam\\
{\tt yujie.xing@outlook.com} \ \ {\tt raquel.fernandez@uva.nl}}

\date{}

\begin{document}
\maketitle
\begin{abstract}
Stylistic variation is critical to render the utterances generated by conversational agents natural and engaging. In this paper, we focus on sequence-to-sequence models for open-domain dialogue response generation and propose a new method to evaluate the extent to which such models are able to generate responses that reflect different personality traits. 
\end{abstract}

\section{Introduction}

The advent of deep learning methods has led to the development of data-driven conversational agents for informal open-domain dialogue \citep[see][for a review]{serbanreview}.  
These chatbot systems model conversation as a sequence-to-sequence (\textsc{Seq2Seq}) problem \citep{seq2seq} and rely on large amounts of unannotated dialogue data for training. We investigate whether such models are able to generate responses that reflect different personality traits. 
We test two kinds of models: The speaker-based model by \citet{persona}, where response generation is conditioned on the individual speaker, and a personality-based model similar to \citet{HerzigEtal-INLG2017}, where generation is conditioned on a personality type. 

Evaluating the output of chatbot systems is remarkably difficult \citep{liu-EtAl:2016:EMNLP}.
To make progress in this direction with regards to personality aspects, we propose a new statistical evaluation method that leverages an existing personality recogniser \citep{oceandetector}, thus avoiding the need for specialised corpora or manual annotations. We adopt the Big Five psychological model of personality \citep{ocean}, also called OCEAN for the initials of the five personality traits considered: Openness, Conscientiousness, Extraversion, Agreeableness,  and Neuroticism. Each of the traits is represented by a scalar value on a scale from 1 to 7.

In the remainder of the paper, we introduce the models we examine and describe our new evaluation method. Our results show that the models are able to generate output that reflects distinct personalities, over a baseline encoding chance personality variation. We conclude with a brief discussion on related work.

\section{Dialogue Generation Models}
\label{sec:models}

The generation models we make use of are standard \seq models consisting of an encoder LSTM, an attention mechanism, and a decoder LSTM \citep{seq2seq,attention0}. 
The model processes context-response pairs, where the context $X=x_1,x_2,\ldots,x_m$ corresponds to the latest utterance(s) in the dialogue and the response $Y=y_1,y_2,\ldots,y_n$ is the utterance to be generated next. 
The probability of the response $Y$ given the context $X$ is predicted as:
\begin{equation}
\textstyle p(Y|X)=\prod_{t=1}^{n}p(y_t|y_1,\ldots,y_{t-1},X)
\end{equation}

\noindent
The attention mechanism by \citet{attention} is used over the hidden states of the encoder LSTM to generate a context vector $c_t$ that determines the relative importance of the words in the context utterance at each decoding step $t$. Then the probability of each word $w_k$ ($k\in|V|$, where $V$ is the vocabulary) to be the next word at step $t$ is predicted with a softmax function:
\begin{equation}
\textstyle  P_t(w_k)=\frac{exp((W)_k\cdot f(c_t,h_t))}{\sum_{k=1}^{|V|} exp((W)_k\cdot f(c_t,h_t))}
\end{equation}
where $h_t$ is the hidden state of the decoder LSTM and $f$ is an activation function. The weights of matrix $W\in\R^{|V|\times d}$ are learned during training, with $d$ being the number of hidden cells.

Both the Speaker Model and the Personality Model we describe below include $4$-layer LSTMs with $1024$ hidden cells per layer.

\subsection{Speaker Model}
Our starting point is the persona-based model by \citet{persona}.\footnote{See \url{http://github.com/jiweil/Neural-} \url{Dialogue-Generation}. We reimplemented the model in PyTorch.\label{ftn:code}}
In this model, each speaker is associated with an embedding $\pmb{v}_s$ learned during training. Whenever a  response by speaker $s$ is encountered during training, the corresponding embedding $\pmb{v}_s$ is inserted into the first hidden layer of the decoder LSTM at each time step (i.e., conditioning each word in the utterance). The hidden states $h_t$ of the decoder LSTM is thus calculated as follows (where $y^*_t$ is the embedding of the response word at time $t$, and $g$ stands for the  LSTM cell operations):
\begin{equation}
\textstyle  h_t=g(h_{t-1},y^*_t,c_{t-1},\pmb{v}_s)
\end{equation}

\noindent
\citet{persona} evaluated their model regarding individual content (factual) consistency. Our goal is to evaluate whether the model preserves individual stylistic aspects related to personality traits.

\subsection{Personality Model}
We modify the Speaker Model to allow for the generation of responses reflecting different personality types. To this end, instead of leveraging speaker embeddings, we estimate the OCEAN scores for each speaker and insert a personality embedding $\pmb{v}_o$ into the first layer of the LSTM decoder.\footnote{The procedure for assigning OCEAN scores to a given speaker is explained in the next section.} 

OCEAN scores are $5$-dimensional vectors $o$, where each dimension ranges from $1$ to $7$. We normalise them to the range $[-1,1]$ and then embed them with a linear layer: $\pmb{v}_o=W_{o}\cdot \frac{o-4}{3}$,
where $W_{o}\in\R^{5\times d}$ is learned during training, thus learning  relationships between OCEAN trait values and properties of the utterances.
Whenever a response with personality traits $o$ is encountered, we insert $\pmb{v}_o$ into the first hidden layer of the decoder LSTM. Thus, the hidden states $h_t$ are now calculated as:
\begin{equation}
  h_t=g(h_{t-1},y^*_t,c_{t-1},\pmb{v}_o)
\end{equation}

\noindent
This version of the model is similar to \citet{HerzigEtal-INLG2017}.\footnote{Our personality model is a modified version of our reimplementation of the code by \citet{persona} (see footnote~\ref{ftn:code}). The code by \citet{HerzigEtal-INLG2017} is not readily available.} The authors focus on the customer service domain and evaluate the model output's style for only two personality traits with human evaluation. In contrast, we deal with open-domain chat and assess all OCEAN traits globally, using the automatic method we describe Section~\ref{sec:eval}.

\section{Experimental Setup}
\label{sec:setup}

\subsection{Dataset}

We use transcripts from two American situation comedy television series: \emph{Friends}\footnote{\url{https://sites.google.com/site/friendstvcorpus/}} and \emph{The Big Bang Theory}.\footnote{\url{https://bigbangtrans.wordpress.com/}}
We consider only those characters who contribute a minimum of 2000 turns, which results in 13 characters (6 from \emph{Friends} and 7 from \emph{The Big Bang Theory}). 
We assign a unique speaker id to each character. In addition, we estimate the personality of each character as follows: for each character, we randomly select 50 samples of 500 utterances each, and estimate the OCEAN scores for each sample using the personality recogniser by \citet{oceandetector}, which exploits linguistic features from `Linguistic Inquiry and Word Count' \citep{pennebaker} and the MRC Psycholinguistic database \citep{mrc}.\footnote{We choose this recogniser because it can estimate numerical scores for each OCEAN trait, instead of binary classifications, and it's open source. For more details, we refer the reader to \citet{oceandetector}.}
We assign to each character the OCEAN score resulting from taking the arithmetic mean of the estimated scores for the corresponding 50 samples. 

We consider every two consecutive turns in a scene to be a context-response pair and annotate each response with either the speaker id or the speaker's OCEAN scores. The resulting dataset contains ${\sim}86$k context-response pairs, of which around 2000 pairs were randomly selected and reserved for validation.

\subsection{Training}

Given the relatively small size of the TV-series dataset, following \citet{persona} we use the OpenSubtitles dataset \citep{osdb} to pre-train the model. OpenSubtitles is a large open-domain repository containing over 50M lines from movie subtitles. Since this data does not include information on which character is the speaker of each line, we simply take each two consecutive lines to be a context-response pair. Due to limitations regarding computational power, we leverage only a subset of the dataset: ${\sim}1.8$M pairs for training and ${\sim}75$k pairs for validation. 

We train a standard \seq model for 15 iterations on the OpenSubtitles training set, until perplexity becomes stable in the validation set. We then initialise the Speaker and Personality models using the parameters learned with OpenSubtitles and train them on the TV-series training set for 30 more iterations, until the perplexity in the corresponding validation set stabilises. 

We use the same settings as \citet{persona} for training: We set the batch size to $128$, the learning rate to $1.0$ (halved after the $6$th iteration), the threshold for clipping gradients to $5$, and the dropout rate to $0.2$. Vocabulary size is $25,000$ and the maximum length of an input sentence is $50$. All parameters (including the speaker embeddings in the Speaker Model) are initialised sampling from the uniform distribution on $[-0.1,0.1]$.

\subsection{Testing}

For testing, we again leverage OpenSubtitles to extract a large subset of ${\sim}2.5$M utterances not present in the training or validation sets. Using each of the utterances in this set as context, we let the trained Speaker and Personality models generate responses for the 13 characters, with Stochastic Greedy Sampling \citep{sgreedy}.
Since general responses are a known problem in neural response generation chatbots \citep{sordoni,serban1,NMI,personality} and our goal is to focus on personality-related stylistic differences, we remove the most frequent 100 responses common to all characters/personalities. After this cleaning step, we end up with ${\sim}700$k responses per character/personality. We refer to the clean set of generated responses as the evaluation set.

\section{Evaluation Method}
\label{sec:eval}

We propose a new evaluation method to measure whether persona-based neural dialogue generation models  are able to produce responses with distinguishable personality traits for different characters and different personality types. 

Using the evaluation set, for each character we randomly select 250 samples of 500 responses and calculate the OCEAN scores for each sample. Recall that the OCEAN scores correspond to 5-dimensional vectors. We label each of these 250 vectors with the corresponding character. This gives us 13 gold classes---one for each character---with 250 datapoints each.
We then use a support vector machine classifier\footnote{We use the SVM implementation in Python's {\tt scikit-learn} library with radial basis function kernel. We tune the regularisation parameter C and use default settings for all other parameters. We tried a range of different algorithms, including $k$-means and agglomerative clustering as well as a multi-layer perceptron classifier, always obtaining the same trends in the results.}
 to test to what extent the OCEAN scores estimated from the generated responses allow us to recover the gold character classes. We compute results using $5$-fold cross-validation (training on 80\% of the set and testing on the remaining 20\% once for each fold). We report average scores over ten iterations of this procedure (i.e., $5\times10$). 

We consider a baseline obtained by randomising the gold character label in the set of generated responses, which indicates the level of performance we may expect by chance. In addition, we use the procedure described above to discriminate between characters using their original (gold) utterances from the transcripts, rather than model-generated responses. This serves as a sanity check for the personality recogniser used to estimate the OCEAN scores---if the recogniser cannot detect personality differences among the characters in the original transcripts, it is not reasonable to expect that the models will be able to generate responses with different personality styles---and provides an upper bound for the performance we can expect to achieve when evaluating generated responses. 

Given that the particular personality recogniser we use  \cite{oceandetector} was not optimised for dialogues from TV-series transcripts, as an additional sanity check we compare its performance on the original (gold) utterances with a bag-of-words (BoW) approach. This allows us to test whether the recogniser may only be detecting trivial patterns of word usage.\footnote{We thank one of the anonymous reviewers for suggesting this additional test.}
We select the top 200 most frequent words over the original utterances as features, without removing words typically considered stop words such as pronouns or discourse markers, since they may be personality indicators. Then we run the same classification procedure using these BoW representations.

\section{Results}
\label{sec:results}

\begin{table*} \centering 
\begin{tabular}{@{}c@{\ \ \ }c@{}}
\begin{tabular}{@{}lccc@{}}\toprule
 \it Friends    & Precision & Recall & F1\\ \midrule
 Baseline     & 0.16 ($\sigma$=.01) & 0.16 ($\sigma$=.01) & 0.16   \\
 Gold          &  0.61 ($\sigma$=.12) & 0.61 ($\sigma$=.16) & 0.61  \\
 Speaker      &  0.32 ($\sigma$=.02) & 0.32 ($\sigma$=.05) & 0.32 \\
 Personality & 0.22 ($\sigma$=.04) & 0.23 ($\sigma$=.09) & 0.23 \\\bottomrule 
\end{tabular}
&
\begin{tabular}{@{\ }l@{}ccc@{}}\toprule
\it Big Bang Theory  & Precision & Recall & F1\\ \midrule
 Baseline     & 0.15 ($\sigma$=.01) & 0.15 ($\sigma$=.02) & 0.15 \\
 Gold          &  0.69 ($\sigma$=.11) & 0.69 ($\sigma$=.16) & 0.69 \\
 Speaker      &  0.46 ($\sigma$=.20) & 0.47 ($\sigma$=.23) & 0.47 \\
 Personality & 0.29 ($\sigma$=.19) & 0.31 ($\sigma$=.24) & 0.30\\\bottomrule 
\end{tabular}
\end{tabular}
\caption{Average scores for 6 characters in {\em Friends} (left) and 7 characters in \emph{The Big Bang Theory} (right)}
\label{tab:results}
\end{table*}

In Table~\ref{tab:results}, we report average F1 score per character (including precision and recall) for the Speaker and the Personality models, as well as the baseline and gold data.  The results for these four conditions are all statistically significantly different from each other.\footnote{Significance is tested with a two-independent-sample $t$-test on the results of 10 iterations, first using Levene's test to assess the equality of variances and then applying Welch's or Student's $t$-test accordingly.}

\subsection{Lower and Upper Bounds}

The first thing to note is that the results on the gold transcripts are higher than the baseline, reaching 61\% F1 score on {\em Friends} and 69\% on {\em The Big Bang Theory}. This indicates that the evaluation method is able to distinguish between the different personalities in the data reasonably well. Apparently, {\em The Big Bang Theory} characters are more distinct from each other than those in {\em Friends}. 

When we use the BoW approach on the gold transcripts instead of the representations by the personality recogniser, we obtain significantly lower results: 23\% F1 score on {\em Friends} and 19\% on {\em The Big Bang Theory}.\footnote{We also run this experiment removing stop words (using the list of English stop words from  {\tt scikit-learn}), obtaining almost identical results: 22\% F1 score on {\em Friends} and 18\% on {\em The Big Bang Theory}.}  
The personality recogniser thus detects patterns that go beyond what can be captured with BoW representations. 

\subsection{Speaker and Personality Models}

We find that the responses generated by the Speaker model display consistent personality variation above baseline, although a significant level of the personality markers found in the original data seems to be lost (32\% vs.~61\% and 47\% vs.~69\%). 
The results obtained for the Personality model are significantly above baseline as well (23\% vs.~16\% and 30\% vs.~15\%). We also see that the personality traits found in the responses generated by the Personality model yield lower distinguishability than those by the Speaker model. This is to be expected, since the Personality model generates responses for a personality type, which should be more varied (and hence less distinguishable) than those by an individual speaker. 

An advantage of the Personality model, however, is that in principle it allows us to generate responses for novel, predefined personalities that have not been seen during training. To test this potential, we create five extreme personality types by setting up the score of one of the OCEAN traits to a high value (6.5) and all remaining four traits to an average value (3.5). We then let the model generate responses to all the utterances in the evaluation set for each of these extreme personalities and evaluate the extent to which the responses differ in style following the same procedure as in the previous experiment. Table~\ref{tab:extreme} shows the results.

\begin{table}[h] \centering
\begin{tabular}{@{}l@{\ \ }l@{\ \ \ }l@{\ \ \ }c@{}}\toprule
                    & Precision & Recall & F1\\ \midrule
 Baseline     & 0.19 & 0.19 & 0.19\\
Average          & 0.53 ($\sigma$=.07) & 0.53 ($\sigma$=.09) & 0.53\\
 {\bf O}pen     & 0.46 & 0.46 & 0.46    \\
 {\bf C}onscientious    & 0.59 & 0.62 & 0.61 \\
  {\bf E}xtravert     & 0.63 & 0.65 & 0.64   \\
 {\bf A}greeable     & 0.53 & 0.50 & 0.51    \\
 {\bf N}eurotic     & 0.44 & 0.42 & 0.43    \\\bottomrule 
\end{tabular}
\caption{Average scores for personality types with high value for different OCEAN personality traits}\label{tab:extreme}
\end{table}

\noindent
We find that the generated responses are distinguishable with 53\% average F1 score. This indicates that the model has learned to generalise beyond the training data. Table~\ref{tab:examples}  shows some examples of generated responses.

\begin{table}[h]
\begin{tabular}{@{}rl@{}}\toprule
Joey \  ({\em Friends}): & \em Oh, of course I love you, baby.\\
Raj ({\em Big Bang}): & \em I don't love you.\\
{\bf O}pen: & \em You are beautiful!\\
{\bf A}greeable: & \em Oh I, I love you too.\\\bottomrule
\end{tabular}
\caption{Responses to {\em Do you love me?}~by the Personality model for personality types of given characters and extreme types not seen during training}
\label{tab:examples}
\end{table}

\section{Related Work and Conclusion}
\label{sec:related}

In recent years, there has been a surge of work on modelling different stylistic aspects, such as politeness and formality, in Natural Language Generation with deep learning methods  \citep[among others,][]{SennrichEtal-2016,HuEtal2017,linguisticstyle,NiuBansal-2018}. 
Regarding generation in dialogue systems, 
besides the two response generation models we have tested, other recent approaches to open-domain dialogue have considered stylistic aspects. For example, \citet{dual-learning} leverage metadata about speakers' personal information, such as age and gender, to condition generation using domain adaptation methods; while  \citet{multi-task-learning}  use multi-task learning to incorporate an autoencoder  that learns the speaker's language style from non-conversational data such as blog posts. The output of these models could also be assessed for personality differences using our method.

More recently, \citet{personagenew} have used the statistical rule-based generator {\sc personage} \citep{personage} to create a synthetic corpus with personality variation within the restaurant domain. They use the data to train and evaluate neural generation models that produce linguistic output given a dialogue act and a set of semantic slots, 
plus different degrees of personality information, and show that the generated output correlates reasonably well with the synthetic data generated by {\sc personage}. 
Our work differs from \citet{personagenew} in several respects: We focus on open-domain chit-chat dialogue, where the input to the model is surface text (rather than semantic representations such as dialogue acts) from naturally occurring dialogue data. Rather than relying on parallel data with systematic personality variation, we exploit a personality recogniser. In this respect, our approach has some similarities to \citet{NiuBansal-2018}, who use a politeness classifier for stylistic dialogue generation. Here we have used the personality recogniser by \citet{oceandetector}, which may not be ideal as it was originally trained on snippets of conversations combined with stream of consciousness essays. Our method, however, is not tied to this particular recogniser---any other personality recogniser that produces numerical scores may be used instead.

We think that the automatic evaluation method we have proposed can be a useful complement to qualitative human evaluation of chatbot models. Our study shows that the models under investigation produce output that retains some stylistic features related to personality, and can learn surface patterns that generalise beyond the training data.

\section*{Acknowledgements}
RF kindly acknowledges funding by the Netherlands Organisation for Scientific Research (NWO), under VIDI grant 276-89-008, {\em Asymmetry in Conversation}.

\bibliography{ref-inlg2018}
\bibliographystyle{acl_natbib}

\end{document}